# Large Scale Generative AI Text Applied to Sports and Music

Aaron Baughman
IBM
Cary NC, USA
baaron@us.ibm.com

Stephen Hammer
IBM
Sandy Springs GA, USA
hammers@us.ibm.com

Rahul Agarwal
IBM
New York NY, USA
rahul.agarwal@ibm.com

Rogerio Feris
MIT-IBM Watson AI Lab
Cambridge MA, USA
rsferis@us.ibm.com

Gozde Akay
IBM
Fredericton, Canada
Gozde@ibm.com

Eduardo Morales
IBM
Coral Gables FL, USA
Eduardo.Morales@ibm.com

Leonid Karlinsky
MIT-IBM Watson AI Lab
Cambridge MA, USA
lenidka@ibm.com

Tony Johnson
IBM
Sandy Springs GA, USA
tonyjohn@us.ibm.com

## ABSTRACT

We address the problem of scaling up the production of media content, including commentary and personalized news stories, for large-scale sports and music events worldwide. Our approach relies on generative AI models to transform a large volume of multimodal data (e.g., videos, articles, real-time scoring feeds, statistics, and fact sheets) into coherent and fluent text. Based on this approach, we introduce, for the first time, an AI commentary system, which was deployed to produce automated narrations for highlight packages at the 2023 US Open, Wimbledon, and Masters tournaments. In the same vein, our solution was extended to create personalized content for ESPN Fantasy Football and stories about music artists for the Grammy awards. These applications were built using a common software architecture achieved a 15x speed improvement with an average Rouge-L of 82.00 and perplexity of 6.6. Our work was successfully deployed at the aforementioned events, supporting 90 million fans around the world with 8 billion page views, continuously pushing the bounds on what is possible at the intersection of sports, entertainment, and AI.

## 1 Introduction

The field of Artificial Intelligence (AI) has been disrupted by the so-called foundation models, i.e., large models (with billions of parameters) pre-trained on massive-scale datasets, which can be further adapted to a variety of downstream tasks with little or no supervision. These foundation models are the basis for Generative AI (Gen-AI) that rearranges neural network building blocks to produce novel content such as text. In the 1980's, expert systems maintained large hand-crafted rules for case-based reasoning. Over time as computing power increased and data grew, task specific and hand-crafted feature representations were input into machine learning techniques. Some of the first techniques included decision trees that automatically learned rules for expert systems. Other techniques emerged such as Support Vector Machines (SVM) and Feed Forward Neural Networks (NN) that were able to solve nonlinear problems. These powerful techniques were foundational for the deep learning movement in the late 2010's. Massive labeled data sets such as WordNet and GPU computing power enabled different neural network topologies to learn task-specific feature representations [1]. Recurrent Neural Networks, Convolutional Neural Networks, Residual Neural Networks, and Long Short-Term Memory Networks are among some of the most popular NN's [2]. Today, self-supervision at scale with massive unlabeled data such as the academic pile and AI supercomputers have created foundational models that are generalizable and adaptable [3, 4]. One such class of foundational models are LLMs.

In 2013, Variational Autoencoders used NN's to encode data into feature vectors and subsequently decoded them back into their original representation [5]. This was one of the first times an encoder and decoder was stacked together. Additional types of NN agents began to work together such as in 2014 with the introduction of Generative Adversarial Networks (GANs) [6]. In this paradigm, one neural agent is a generator and produces content while a second neural agent is a discriminator that discerns between real world samples and generated samples. In 2017, a seminal paper called, "Attention is All You Need", depicts ways stacks of encoders and decoders can be linked together with attention heads [7]. This enables networks to determine the most important parts of the input. Now in the 2020's, many different types of LLMs have become available such as T5, BERT, GPT, PaLM, Granite, and BLOOM [8, 9, 10, 11, 12, 13]. These networks are foundational LLMs that are trained on different tasks. The techniques of Prompt Engineering and Instruction Tuning was introduced by FLAN models [14]. However, LLMs are becoming challenging with their large size.

Through an approximate comparison, the human brain has an estimated 100 billion neurons and many LLMs have orders of magnitudes more parameters [15]. The number of parameters is growing at an exponential and unstainable pace. For example, T5-large started with 770 million parameters [8]. Different versions of IBM Granite have up to 13 billion neurons with Llama 2 having a 70B flavor [12, 16]. GPT 3 has 175 billion parameters with GPT 4 having a potential number of 100 trillion parameters [10]. Just to train IBM Granite 13 billion, 153,074 kWh is required [12]. To run these LLMs for production jobs, specialized GPU machines with at least 64 GB of memory are required. Today, research is ongoing around the field of model distillation such that large models can be shrunk without performance compromises [17].

Within our paper, we tap into the capabilities of Gen-AI, with a focus on Large Language Models (LLMs), to address the problem of scaling up the production of media content, such as commentary and personalized news content for some of the most prominent sports and entertainment events in the world. We began by investigating smaller encoder and decoder model architectures that could be run on private bare metal machines. We used T5 transformer models within the Masters golf tournament to translate golf ontologies into sentences that are subsequently paraphrased many times. Next, we selected a larger model called IBM Sandstone 3 billion to fine tune to domain of tennis [18]. The techniques such as PEFT Low-Rank Adaptation (LoRA) and Alibi enabled us to create 4 model deltas for tennis scenes that translated structured data in the form of JSON to text for commentary [19, 20]. With large models, model hallucination was a challenge, which resulted in us changing to decoder architectures with few shot learning. For ESPN Fantasy Football, the Llama 2 7 billion model with over 1 trillion tokens created fill in the blank sentences that could be personalized by consumer facing applications. At the 2024 GRAMMYs, Llama 2 70 billion with few shot learning and RAG created novel text elements about music artists that are integrated into live video feeds of the show and published on social media as multi-media assets [21]. This paper depicts our journey of applying, adapting, and deploying Gen-AI to:

- The Masters Golf Tournament: To provide commentary on every shot from every player from every hole and round within the style of fact-based golf commentary delivered through audio and text captions that were paired with the associated video clips in near real time.
- The Wimbledon Championships and US Open Tennis: To commentate match start, match end, and set end points within a colorful commentary style delivered within an audio and text captioning format, synchronized with the associated video clips of the match.
- ESPN Fantasy Football: To generate personalized evidence-based sentences that describe the rationale for fantasy football player performance grades.
- GRAMMY Awards: To generate timely headlines, bullet points, witty text, and short form articles about music artists delivered as insights overlayed on video streams and packaged video assets to share on social media platforms.

## 2 Related Works

The field of AI is rapidly adopting neural network-based models that are generalizable across domains with the ability to adapt to many downstream tasks. These kinds of algorithms are called foundation models, which is built from massively labeled data, large compute clusters, and deep learning elements [3, 4, 22, 5]. Foundational models are broken into classes of encoders, decoders, and encoder-decoders. Encoder models experienced explosive growth with the popularity of BERT [9]. Given an alphabet or tokenized input, these types of models create a high dimensional representation of input feature vectors that can later be used for classification. Encoder-decoder models are sequence to sequence algorithms that can transform an input into a desired output. Models such as T5 popularized this class of algorithms [8]. With the introduction of GPT-3, the decoder only architecture that changes the representation of an input alphabet into a new output became very popular [23]. Over time, the encoder only architecture began to fade with popularity as the encoder-decoder and decoder architectures showed good performance [24].

The union of decoder, decoder-encoder, and encoder styles of algorithms have created natural language models called Large Language Models (LLM) that can reason with text. Many of the supported use cases for LLMs include Natural Language Generation, Natural Language Understanding, Chain of Thought Reasoning, and Knowledge-based tasks [24, 25]. The commercialization of LLMs have been supported widely from companies such as IBM, Meta, Google, Microsoft, and AWS [17, 16, 14, 26, 27]. Opensource communities and frameworks such as Hugging Face and Pytorch that can run on hosted cloud components are increasing the adoption rate of LLMs within industry [28, 29]. Within our work, we implement and empirically evaluate decoders and encode-decoder architectures on both public and private cloud systems. Over time, the methodology in which we trained and adapted LLMs changed as prompting techniques matured.

### 2.1 Fine Tuning

The ongoing trend shows that the more parameters a LLM has the more flexibility, it has to be able to learn different tasks [23]. However, larger models require high end GPU's to prompt or fine tune with exemplars and they are prone to hallucination at inference time [.]. Techniques such as PEFT reduces the number of free parameters that are trained by freezing layers [19]. A few types of PEFT techniques are widely available. Delta tuning enables the freezing of parameters within regions of a LLM to reduce the number of free parameters [.]. The adapter tuning method inserts modules between layers and focuses training only on those parameters [30]. LoRA creates low-rank matrices through matrix decomposition to reduce the number of trainable parameters [31]. The prefix-tuning focuses on task specific vector features while freezing all other parameters [32]. By using the LoRA PEFT method when fine-tuning our LLMs, we were able to use our own bare metal machines described in Section 4.2. In addition, we used Alibi to extend the number of tokens during inference time [20].

Techniques such as zero shot, one shot, and few shot learning have reduced the need for expensive fine tuning. Larger models are very effective when given input and output examples to influence the output of the model [23]. On 9 different tasks, work by Tom Brown showed that few-shot learning with large models generally performed well without the need for fine tuning. This type of in context learning with a few examples help LLMs produce the correct output for unseen input [33]. A type of example generation called self-instruction uses a LLM to produce instruction input and output samples and then filters out invalid responses [34]. These samples are used as examples for few shot learning [.]. To further help LLMs produce relevant and timely output, a technique called Retrieval-Augmented Generation (RAG) queries external sources and adds content to a context block within a prompt for input into a LLM [21]. The progression from fine-tuning to few shot learning with RAG enhanced our production deployments within sports and entertainment.

### 2.2 Large Scale LLMs

Both running and training LLMs require a lot of compute resources and data. The academic pile is a large set of 800 GiB data from 22 sources [3]. For example, the pile contains data from PubMed Central, arXiv, GitHub, US Patent and Trademark Office, YouTube and etc. [.]. Many companies augment this data pile with enterprise and proprietary data [12]. These large piles accelerate the development and experimentation of LLMs. AI supercomputing clusters are required to train these models with data piles of terabytes in size and many LLMs having over a billion parameters.

A large investment is generally required to build and train a LLM. IBM created the Vela AI supercomputer GPU cluster [4]. The cluster has powerful compute nodes with 8 80GB A100GB GPU's, 1.5TB of DRAM, 96 vCPUs, and four 3.2 TB NVMe drives [.]. The nodes are connected with a 100G link to each Virtual Machine that provides compute. On this cluster, IBM's Granite 13B v2 model used 256 A100 GPUs for 1152 hours and 120 TFLOPs that consumed 153074 kWh [12]. Other companies such as AWS have their own infrastructure and frameworks such as MiCS for gigantic model training [27]. NVIDIA has developed large scale clusters that support models with over 1 trillion parameters at 502 petaFLOP/s on 3072 GPUs [36]. Microsoft created the Varuna cluster that trained models with 200 billion parameters on VM's with shared GPU's [36]. With the increasingly number of different compute clusters that use AI accelerators, groups are beginning to study the comparative characteristics of each. For example, Emani's work compares Nvidia A100, Samba Nova, Cerebras CS-2, Graphcore Bow-Pod64, Habana Gaudi2, and AMD MI250. Many of the gains for a particular AI accelerator is tied to the parallelization capability of an LLM [37].

## 2.3 Text Applications

The application of LLMs to user experiences is just starting and gaining adoption within the industry. One of the most popular and large-scale systems is Google's generative AI in search (SGE) [38]. Though there is not a lot of scientific literature about the feature, we can speculate the system uses an iteration or technique from PaLM, a 540 billion parameter transformer language model [11]. Another consumer self-serve feature is OpenAI's chatgpt-4 interface [10]. A chat-based interface enables user's to pose questions and to have conversations with a generative agent [.]. There is a lack of academic literature that discusses generative AI systems that must handle up to 30,000 rps such as our system discussed in Section 4.2.

## 3 Live Events and Data

As the official AI and Hybrid Cloud provider for the GRAMMYs, Masters, Wimbledon, US Open, and ESPN Fantasy Football, we are creating generative AI fan experiences for sports and music fans around the world. Each of our properties have a unique set of needs, challenges, priorities and goals which requires us to continuously innovate for different use cases. For example, in golf, we wanted to create shot by shot AI generated commentary that is factual, concise, and monotone to match the style of the sport. We used golf shot, score, player, schedule, ball positioning, hole information, leader board, and player performance distribution data from 2016 to the present. An accumulated 169,614 shot elements were available to fine tune T5 models. In total 245 unique tuples were generated and annotated with sentence ground truth that was agreed upon by Augusta National Golf Club and IBM golf Subject Mater Experts (SMEs). Next, each of the 245 sentences were paraphrased 5 times for a total of 1225 ground truth sentence templates. During training, each of the templated sentences were filled with values that followed golf rules. To test the scale of the system, the entire previous year's Masters golf tournament was replayed in real time.

Within tennis, we were tasked to create AI-generated commentary that sounded like a human broadcast commentator (with color and personality), as the use case was adding commentary to match highlight videos. A total of 2580 match start, match end, and set end shot scenes was available for fine tuning across both Wimbledon and the US Open with 10% held out for validation. Tennis JSON and sentence pairs were annotated with approved style from either American or British English sentences. A total of 141 templates were automatically filled in to create 30,000 exemplars for fine tuning an IBM Sandstone 3B model. Each exemplar-based sentence was generated satisfying tennis rule conditions.

At ESPN, we designed a high scale system that generated fill in the blank sentences or slot fillers that could be personalized for each of the 12 million users to describe why each football player was given their grade. The system was shown fill in the blank examples for each feature and percentile level. For example, 20 examples were generated for 13 statistics mined from data distributions back to and including the 2015 season. The percentile score for a particular football player statistic such as projected weekly score controlled the style of the sentence. A total of 1211 template sentences were used to generate 20 examples for few shot learning on each validation test. 1014 samples that were assembled by football SMEs were used for validation.

For the GRAMMYs, the generative system creates headlines, witty text, bullets and short form narratives. A selected top 136 artists based on this year's nominations are used for testing. Depending on the type of generated content, the system pulls from current Recording Academy articles, headlines, statistics, artist biographies, and Wikipedia. The most relevant data is included within RAG for the generation of text. Prompt Engineering is used to refine our outputs by adding negative prompts and other safeguards to the system as well as modification on the length and the tone of the text. The perplexity of each type of content was measured for each of the 136 artists and reviewed by a team of music SMEs before the content was published to millions of music fans around the world.

## 4 Generative AI Architecture

A generative text system depicted in Figure 1 supports golf, tennis, American football, and music content creation. Golf and tennis text generation is triggered by player results – such as completing a shot in golf or winning a set in tennis. Within golf, every hole completion or score change is sent to an Event Streams Kafka partitioned topic. Similarly, within tennis, score events such as set end and match start is posted to another set of Kafka partitions. A Language Generation algorithm pulls messages from the topics and distributes the work to other services. Additional messages with LLM prompts are routed to one of T5-Large, IBM Sandstone 3B, Llama 2 7B, or Llama 2 70B. The results of the LLMs are saved within a document-based database, Cloudant, and an object database, Cloud Object Storage (COS). Consumer devices access the textual components in the form of auditory content, captions, or text through a Content Delivery Network (CDN) with COS as origin. Periodically, scores and statistics are updated, which require the generative content pieces to be updated. Update messages are sent to the Kafka topics for processing as new events.

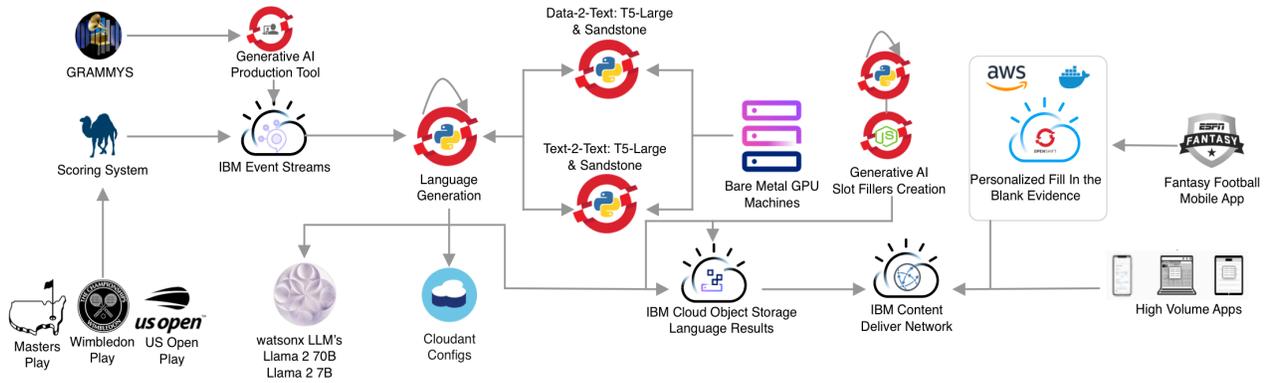

Figure 1: Generative Architecture for the Masters, US Open, Wimbledon, GRAMMYS, and ESPN Fantasy Football

For direct consumer facing requests such as ESPN Fantasy Football, a novel process creates slot filler content. Two applications written in Python and Node pull and create JSON files with slot sentence dictionaries based on statistical percentiles and type. A batch process is started every hour that assembles a prompt with few shot learning and posts for every football statistical type to a Llama 2 7B model to generate filler sentences. A total of 1350 slot filler sentences are created and transformed into JSON files for storage within COS for distribution across the world within a CDN.

As user's request the need to understand the grades of fantasy football players, an application fills in slot values for personalized evidence-based sentences. The application receives a payload that describes a user's team roster, league, rules, and team weaknesses. This information is used to create user specific information such as the amount a player could help an owner's team. The filler sentences are retrieved from the CDN based on the statistical type and percentile that best describes the rationale of a player's grade. Next, the application fills in the slot values and returns the personalized evidence-based sentences for rendering on the User Experience (UX). As the batch job runs, the slot fillers change to adapt to changing language and style within fantasy football.

| Sport | Input | Ground Truth Output |
|---|---|---|
| Golf | On_hole_number_3,_Golf Player One \| is attempting\| his putt for par | On hole number 3, Golf Player One is attempting his putt for a par. |
| Tennis | "player_one":{…}, "player_two":{…}, "draw":{…}, "head_to_head":{…} | Player One ranked 46$^{th}$ in the world, will play against Player Two, who looks to take home her 6$^{th}$ win against Player Two |
| ESPN Fantasy Football | {FIRST_NAME':'Name1', 'LAST_NAME':'Name2', 'POSITION': 'Running Back', 'NEED_PERCENTAGE':0.9, 'NEED_PERCENTILE':92} | Big bump in the {position} position by acquiring {last_name}. \n\nDone |
| GRAMMYS | Instruction Preamble  Write 3 summary sentences separated by a * about Artist Name without any numbers or stats and a strict 150 character limit. Do not repeat the instruction. | * Artist Name is a talented singer-songwriter known for her catchy pop tunes and heartfelt lyrics.  * She has released several successful albums and singles, |
| | RAG Context | and has won numerous awards for her music.  *Her songs often focus on themes of love, self-empowerment, and friendship, resonating with fans of all ages. |

Table 1: Input and output pairs for Golf, Tennis, ESPN Fantasy Football and the GRAMMYs.

## 4.1 Pre-Processing

For each of the types of score-based or consumer-based messages, data is organized for rapid retrieval. A graph-based ontology Resource Description Framework (RDF) is created to model relationships of entities. For example, within golf, nodes that represent golf are organized to increasing granularity of tournament, round, and hole. An RDF query language is executed on the always updating graph structure. An example query is depicted in Figure 2.

```
select ?s WHERE{?s rdfs:subClassOf hole:HOLE.
FILTER (?s=http://masters.ontology.ai/hole_5)} limit 1
```

Figure 2: RDF query for golf

A series of data congruency checks in the form of rules, ensure data congruency. For example, within golf, the track feed could update before the score feed. When this happens, the data extracted from the live feeds will be inconsistent. Similarly, data can be missing. Within ball tracking, if the golf ball laser tracking system cannot acquire the location of the ball, humans enter the information. The delay can cause a data gap. Within a third case, the information could be wrong within a feed. A tennis court tracking device or an incorrect score could be entered into our live scoring system. This is apparent when a score does not follow legal rules or conflicts with another feed. In all cases, the system will wait 5 seconds and check again for data consistency before requeuing the message back onto Kafka for reprocessing.

Even with preprocessing, many scenes need to be updated very quickly to expand a golf shot or to include breaking news about a music artist. A separate generative AI workflow that is shielded from the large-scale traffic of a live event is optimized to replace older generative content. Update messages are sent to a specific set

**Football Prompt Template**

{{#instruction_prefix}}\n{{instruction_prefix}} {{instruction}}\n{{/instruction_prefix}}\n{{^instruction_prefix}}\n{{/instruction_prefix}}\n{{#input_prefix}}\n{{input_prefix}} {{input}}\n{{/input_prefix}}\n{{^input_prefix}}\n{{input}}\n{{/input_prefix}}\n{{#examples_prefix}}\n{{examples_prefix}} {{examples}}\n{{/examples_prefix}}\n{{^examples_prefix}}\n{{examples}}\n{{/examples_prefix}}\n{{output_prefix}}"

**Golf/Tennis Prompt Template**

{{#input_prefix}}\n{{input_prefix}} {{input}}\n{{/input_prefix}}\n{{^input_prefix}}\n{{input}}\n{{/input_prefix}}

**Golf/Tennis Prompt Creation**

input: Golf Player is playing On hole 9 AND he is hitting From the Pine Straw

**Football Prompt Creation**

instruction: Create a bullet point about next game projection.

examples:

input: {'FIRST_NAME': 'First 'Name, 'LAST_NAME': 'Last 'Name, 'HAD_IR': False, 'HAD_BYE': False, 'HAD_SUSPENSION': False, 'OPPONENT_NAME': 'Atlanta Falcons', 'NEXT_GM_PROJ': 15.55, 'NEXT_GM_PROJ_PERCENTILE': 82}
output: {last_name} who will play against the {opponent} is projected to score an outstanding {projection_points} points.
\n\nDone
…

**GRAMMYS Template**

{{#input_prefix}}\n{{input_prefix}} {{input}}\n{{/input_prefix}}\n{{^input_prefix}}\n{{input}}\n{{/input_prefix}}\n\n{{#context_prefix}}\n{{context_prefix}} {{context}}\n{{/context_prefix}}\n{{^context_prefix}}\n{{context}}\n{{/context_prefix}}\n\n{{#avoid_topic_prefix}}\n{{avoid_topic_prefix}} {{avoid_topic}}\n{{/avoid_topic_prefix}}\n{{^avoid_topic_prefix}}\n{{avoid_topic}}\n{{/avoid_topic_prefix}}\n\n{{output_prefix}}

**GRAMMYS Prompt Creation**

input: Clean Preamble Phrase. Write 3 sentences seperated by a * about Male Artist Name with masculine pronouns without any numbers or stats and a strict 150 characters limit. Do not repeat the instruction.

context: {sentences queried about the topic}

avoid topic: violence

Figure 3: Prompt templates applied to Golf, Tennis, Football, and the GRAMMYS.

of Kafka partitions to fast-track processing. The resulting generative content pieces are generated and uploaded to COS and Cloudant. Finally, a CDN cache purge request is issued to the cloud components to refresh the data for consumer devices. Equation 1 depicts the creation of a pre-processed data vector, $\bar{x}$, and prompt, p, into well-formed data, $\bar{x}'$, and a corrected prompt p'.

$$\bar{x}', p' = \text{pre}(\bar{x}, p) \quad (1)$$

## 4.2 Large Language Models

Throughout our work, we have progressively implemented larger generative models to create text at scale. After experimentation with a desire for a finely controllable model, we started with the 770 million parameter T5-Large model. This transformer model is very descriptive when fine-tuned with exemplars with little risk of hallucination. Two T5-Large models were trained and nested together. The first one was used to transform tuples into sentences while the second one paraphrased a single sentence into 5 sentences. Next, we used the IBM Sandstone 3 billion parameter transformer. Although this model had 20% more risk of hallucination, the sentence variation was much higher giving the AI commentary more personality. A single IBM Sandstone model could transform a tuple into 5 sentences. Next, we switched to decoder only type models. We began using the Llama 2 7 billion parameter model. The risk of model hallucination with few shot learning was less than 5% higher than the descriptive T5-Large models. The examples provided within the prompt guide the model with the conversion of different types of input into specific output formats. An open domain task of generating free text with an even larger model, Llama 2 70 billion parameter model, was effective with the RAG technique. The retrieved content about a topic is placed as context in a prompt that is used as anchor information during text generation. In general, prompt engineering decisions were critical for the creation of high-quality text. Equation 2 shows an engineered prompt, $p''$, created from filling in prompt templates.

$$p'' = \text{pre}(\bar{x}', p') \quad (2)$$
$$t_{raw} = \text{LLM}(p'') \quad (3)$$

Figure 3 depicts a mustache style of prompt templates. The golf and tennis templates are basic transformer style templates. The input prefix is defined so that the tuple is inserted into the prompt. The football prompt template is an example of few shot learning. Through experimentation as seen in section 5, 20 examples inserted into the prompt template produced the best results. The LLM was able to infer how to convert the JSON representation of football data into fill-in-the-blank sentences. The GRAMMYs template depicts the use of RAG. The context block contains data retrieved from a vector database, Cloudant search, and IBM Watson Discovery search. The text in the context focuses the LLMs output to be relevant with outside knowledge. The avoid topic section indicates the concepts to eliminate from the generated text. The instruction in the prompt indicates to the LLM what gender pronouns to include within the generated text. All the prompts have a preamble which tells each LLM not to include text about Hateful and Profane (HAP) content.

The hosted T5-Large and IBM Sandstone 3B ran on 5 bare metal machines with a single NVIDIA Tesla V100 GPU that had 32 GB of RAM. In the best case, this GPU configuration could provide an inference response time in a few seconds and take hours to train. We were able to utilize bf16 to train the T5-Large model and included PEFT LoRA to optimize the training of the IBM Sandstone 3B model. However, when we moved to the Llama 2 70B model, the GPU was not sufficient. In general, about 1.4GB of memory is required to load 1B parameters, which means 98GB of memory would be required. We used the WatsonX platform that hosted the Llama 2 models. Equation 3 depicts the application of the LLM to an engineered prompt to produce raw text, $t_{raw}$.

When we looked at our highest rps scale use case at ESPN Fantasy Football, the LLM system would need to support 30,000 rps with 834 parallel Tesla V100's. We created a batch system that used a LLM to generate fill-in-the-blank sentences. A consumer facing application was scaled out on 120 OpenShift PODs over two regions to absorb the 30,000 rps traffic. Each region has 3 worker nodes with 16 vCPU's and 64 GB of memory. The application that runs in the PODs loads the pre-generated fill-in-the-blank sentences and personalizes the value fillers before returning to the user.

Our hosted LLMs such as Llama 2 70B have a reduction of random text by minimizing creativity with a low temperature that falls between 0 and 0.2. The top_k and top_p values are kept at conservative values such as 10 and 1, respectively. The prescriptive hyperparameters enable us to minimize humans in the loop for text review. For example, golf and football text generation was fully automated without any humans whereas tennis and the GRAMMYs had 5 human reviewers.

### 4.3 Post-Processing

Each of the models and generative workflows we built and deployed had different challenges. For golf within 2023, the T5-Large transformers were prone to misspell names or to mix up golfers. The error rate when contrasted to ground truth of our best performing model was at 16%. A golf name spell checker along with clues within the sentence such as nation of citizenship, player rank, and schedule helped to disambiguate names with close word edit distance. At Wimbledon and the US Open in 2023, the IBM Sandstone 3B model hallucinated 20 times out of a 100 with a moderate temperature of 1 and beam number of 5. A series of regular expression fact checkers classified the type of sentence while parsing out generated statistics. The statistics were compared to streaming ground truth scores, schedule, draws, and head-to-head statistics. At other times, non-conforming vocabulary words also provided indicators of a model hallucination. After detection and shown in Equation 4, the hallucination was mitigated by correcting the words, name, or statistic to produce text, t.

$$t = \text{post}(t_{raw}) \quad (4)$$

The effort of post processing was moved to example synthesis during few shot learning. This was the case at ESPN Fantasy Football. A minimal number of post processing rules were required to fix less than 1% of generated fill-in-the-blank sentences using the Llama 2 7B model with a moderate temperature of 1 and top k of 50. For the GRAMMYs, the use of RAG greatly enhanced the factualness and timeliness of artist headlines, bullet points, witty text, and summaries. Prompt engineering with the Llama 2 70B was of paramount importance to eliminate irrelevant, old, or offending content.

When creating content about people such as artists, we ensured we used the correct pronouns. For example, some artists go by feminine, masculine, or neutral pronouns. Through our prompts, we instructed models to use a class of pronouns. The output of the large decoders changed the text to match a person's desired pronoun use. To correct any textual errors, a set of regular expressions were written to ensure the selected pronouns matched the desired gender. At the GRAMMYs, we had a human in the loop to check the content for both style and pronouns before it was included within an insight or social media story.

## 5 Results

Throughout our applied work to live sports and entertainment events around the world, we selected several industry level objective functions to measure the quality of our generated text. The first one we used for short sentences was word edit distance or the Levenshtein distance metric. Equation 5 shows the formulation of the word edit distance where tail(t) is text t without the first letter and |t| is the length of the text. The lower the score, the better. However, in our work, we standardized the score such that the closer the value is to one, the better.

$$\text{lev}(t_1, t_2) = \begin{cases} 1 + \min \begin{cases} \text{lev}(\text{tail}(t_1), t_2) \\ \text{lev}(t_1, \text{tail}(t_2)) \\ \text{lev}(\text{tail}(t_1), \text{tail}(t_2)) \end{cases} \text{otherwise,} \\ \text{lev}(\text{tail}(t_1), \text{tail}(t_2)) \text{ if } t_1[0] == t_2[0], \\ |t_1| \text{ if } |t_2| == 0, \\ |t_2| \text{ if } |t_1| == 0 \end{cases} \quad (5)$$

To compliment the standardized word edit distance, we also use $\text{Rouge} - N$ measures. The $N$ represents the n-gram overlap between the system generated text and the ground truth. We used both $\text{Rouge} - L$ and $\text{Rouge} - 2$ for best coverage. In Equations 6, 7, and 8, the notation t is the generated text, g is the ground truth, and ngram is the gram overlap selection.

$$\text{Rouge} - N_{recall} = \frac{\text{ngram}(t,g)}{|g|} \quad (6)$$

$$\text{Rouge} - N_{precision} = \frac{\text{ngram}(t,g)}{|t|} \quad (7)$$

$$\text{Rouge} - N = 2 * \frac{\text{Rouge}-N_{recall} * \text{Rouge}-N_{precision}}{\text{Rouge}-N_{recall} + \text{Rouge}-N_{precision}} \quad (8)$$

Some of our use cases did not have a clear ground truth. One such case where the generation of text is very open ended is at the GRAMMYS. In this case, we measured the perplexity or the quality of the model with respect to being surprised by the output. The perplexity measures the probability that a word is generated given the probability of all other words shown in Equation 9.

$$\text{Perp}(t) = e^{-\frac{1}{t}\sum_i^j \log p_\theta(t_i|t_{<i})} \quad (9)$$

The generated text, t, is indexed by a word, i, such that the quality of the model can be measured.

In real world applications, system performance and tradeoffs are just as important as empirical objective metrics depicted in equations 5, 8, and 9. When a quick decision for deployment was required for several of the models, we created an "Approach Score Card". We scored models based on ordinal numbers [1,10] where 1 is the lowest and 10 is the highest. Factors such as infrastructure, number of models, etc. are combined. Section 5.2 goes into additional detail.

### 5.1 Text Measures

The following tables summarize our objective measures we discussed in Section 5. We started with the 2023 Masters. Since this was the first large scale consumer facing project with generative text, we wanted to finely guide the text output. As a result, we picked the T5-Large model. The transformer style model was small enough for us to fine tune the encoder and decoder layers to accept tuple input that would be output as fact-based commentary style sentences. In 16 experiments we boosted several underperforming cases such as rare water shots. The best T5-large models that averaged both tuple to text and the paraphrasing of text had a standardized word edit distance of 80.68, Rouge-L of 98.45 and Rouge-2 of 97.11. With the combination of post processors discussed in Section 4.3, we deployed the system at the 2023 Masters and processed over 40,000 near real-time golf scenes without any human review. Going into the 2024 Masters, we are evaluating few shot learning with hosted models on IBM WatsonX that are outperforming our best 2023 T5 model. The Llama 2 7B model is our top performer followed very closely by IBM Granite Instruct 13B.

| Golf | Granite Instruct 13B | Granite Chat 13B | Llama 2 7B | 2023 Best T5 Models |
|---|---|---|---|---|
| Std Word Edit Distance | 97.89 | 95.61 | 98.35 | 80.68 |
| Rouge-L | 98.87 | 97.39 | 99.12 | 98.45 |
| Rouge-2 | 97.88 | 95.92 | 99.84 | 97.11 |

Table 2: Golf model evaluation

The generated golf text can be seen within Figure 4. The caption shows the end results of converted a golfer's post shot tuple into a sentence that describes the stroke, location, and the distance left for the ball to travel into the hole. The converted sentence is then fed into IBM's text to speech service to produce an audio file. The corresponding audio file and captions are paired with the video clip for every shot. User can listen to pre and post shot commentary generated by AI for all 20,000 golf shots during the tournament.

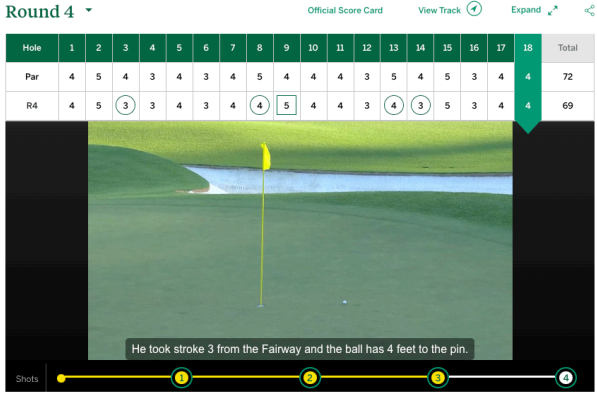

Figure 4: The Masters AI commentary captions depicted within a shot highlight.

Tennis was a different type of use case than golf. Not only is it a completely different sport, but we wanted more personality and variation within the generated text. The highly guided workflow of the T5-Large model did not provide enough text diversity for tennis. Through several experiments, we decided to use an IBM Sandstone 3B model that had a similar topology as T5-Large but could produce a bit more text variety. Table 2 shows the Rouge-N scores for the Sandstone 3B model. We created commentary for 4 types of tennis scenes: the start and end of the match, a set point within the match, and the set point to conclude the match. Overall, the metrics were good, however, some of the output of the model created hallucinated facts. As a result, 5 humans reviewed the output of the model to make corrections before publication to the consumer facing experience: AI Commentary for AI Highlight packages.

| 2023 Tennis IBM Sandstone 3B | Match Start Scene | Match End Scene | Set Point End Scene | Match Point End Scene |
|---|---|---|---|---|
| Rouge-L | 76.4 | 86.8 | 71.9 | 74.7 |
| Rouge-2 | 69.1 | 84.2 | 58.3 | 71.6 |
| Rouge-1 | 80.3 | 90.7 | 76.4 | 84.0 |

Table 3: Tennis model evaluation

The highlight packages that are automatically generated include the gen-AI text. An example caption is depicted within Figure 5. Each of the highlight package's match start, set end point, and match end has corresponding captions and speech. The generated text allows new avenues for fans to interact with the content by providing emotional context of the tennis play.

Within football, we wanted to continue the variety of text generation to explain why and how we graded football players within ESPN Fantasy Football. These player grades were personalized to every user based on their current team composition. We began experimenting with few shot learning around a new class of LLMs: decoders. Table 4 shows the results of standardized word edit distances and rouge scores for larger Llama 2 and IBM Granite models. For this task, we wanted to create fill-in-the-blank sentences that were later personalized for each fantasy football team owner. We did not fine tune any of the models and provided 20 input and output examples to show the model what type of text to produce. The Llama 2 7B model was our preferred choice for this task. The output of the model did not require any human review before their availability within the consumer facing application.

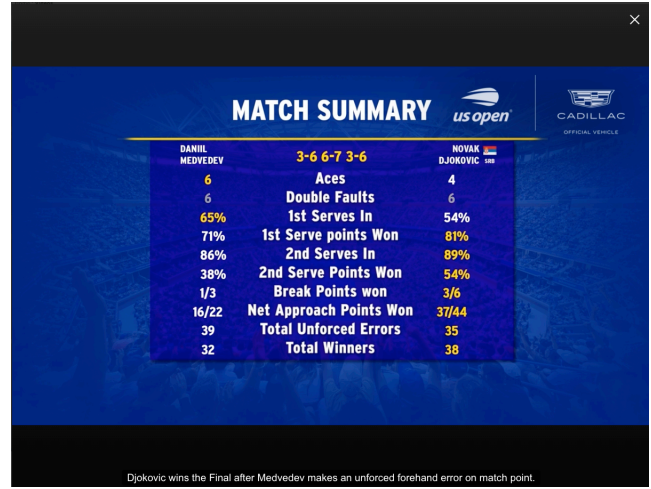

Figure 5: 2023 US Open men's singles championship winning shot narrated by generative AI example. The caption is available on the bottom of the video, which is rendered in parallel to a synthesized commentator voice.

| 2024 Football | Llama 2 7B | Llama 2 70B | Granite Instruct 13B | Granite Chat 13B |
|---|---|---|---|---|
| 1-Word Edit Distance | 89.5 | 88.8 | 43.7 | 32.0 |
| Rouge-L | 86.8 | 86.0 | 67.0 | 39.1 |
| Rouge-2 | 83.1 | 81.7 | 58.0 | 27.2 |

Table 4: Football model evaluation

At the GRAMMYs, we had two open-ended problems that we solved that with a gen-AI user experience (UX) shown in Figures 6 and 7. As artists walk down the red carpet, we want to create generative textual insights that provide context about the artist. From our experiences with golf, tennis, and football, we wanted to pick a model that could be highly creative while being able to contain model hallucination. The optimization of the perplexity metric helped us to select the Llama 2 70B model. We picked model temperatures in the upper 80% of values and increased the top k selections to 100. To help guide the model towards timely and factually information, the RAG method enabled the model to focus on information we retrieved from the Recording Academy ecosystem. In another use case, we created social media assets that pulled in the artist generated text into templatized videos. These videos are shared across IBM and Recording Academy social media channels. We had 5 humans that corrected and manually published the insights for both use cases. This is a self-service generative AI tool and can be used not only for the live show, but throughout the year to help scale and automate the production of GRAMMY content for the Recording Academy Editorial team.

The results for this project are broken into two types: free and categorical. When users create a generative content for an artist, they enter in a category such as GRAMMY Achievements, Musical Influences, Activism, or etc. to bias the retrieved and generated content's source of relevancy to both the artist and category. The content will be focused around the category a user has selected. The free type enables the LLMs and RAG algorithms to select and generate content that has the highest relevancy towards the artist. The two distinctions enable users to create text for the livestream experience and social media assets that are potentially directed to late breaking news or the most interesting aspect about an artist.

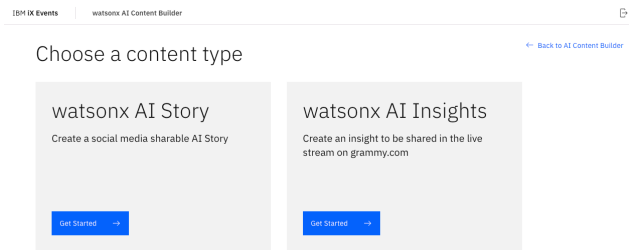

Figure 6: GRAMMYs gen-AI UX to produce AI Story and AI Insights.

| Music | Best 2024 GRAMMYS Model – Llama 2 70B |
|---|---|
| Free Perplexity | 5.34 |
| Categorical Perplexity | 7.86 |

Table 5: Music (GRAMMYS) model evaluation

## 5.2 System Measures

Through our 6 large live scale events, we collected operational and engineering information to help guide us on future decisions. Figure 7 shows the features that were important to our decision making. To keep our workflows simple, we wanted to minimize the number of LLMs, use hosted hardware, reduce post processing and hallucination, and optimize community support. Overall, the two models with the highest operationalize score for future events are the IBM Granite and Meta Llama 2 classes of models. The IBM Granite models scored the best with a total score of 9 because the model is a managed service on IBM watsonx, can handle multiple tasks, needs a low amount of post processing, and is actively supported and improving. The Llama 2 models scored slightly worse in hallucination possibilities based on our detectors. With the worst score, the T5 series models take significantly more effort to manage and deploy than either the Llama 2 and IBM Granite Models. As new models evolve and are released into market, we will continually update our decision matrix.

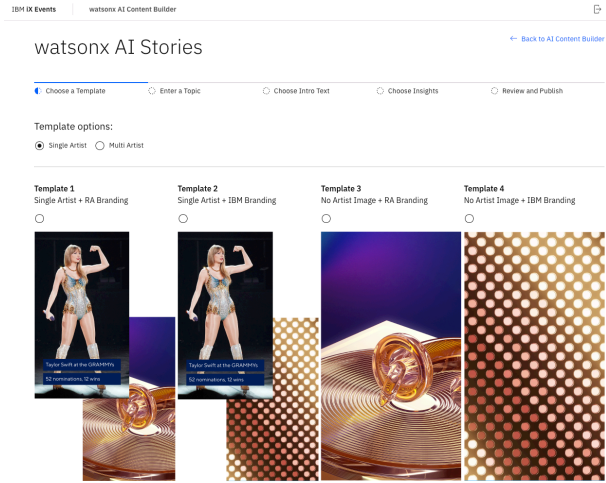

Figure 7: Operational model evaluation for practical deployment and management.

Figure 8: GRAMMYS AI Story UX that supports the production of social media videos.

## 6 Future Work

For future live events within the sports and entertainment industry, we are investigating the development of generative multimedia workflows. We would like to combine specialized content creators such as LLMs with diffusion models, Large Vision Models (LVM), and Generative Adversarial Networks (GANs). The next step towards general content producers will include models that jointly create text, sound, images, and video from a plurality of input. The next generation of our human centered experiences will showcase generative multimedia touch points. Further, we would like to extend model distillation work to multimedia and to support multiple languages [17].

Throughout our work, we spent a lot of effort writing post processors to correct hallucination errors. Many of our decisions were deliberately made to minimize the risk of incorrect narration of text disseminated to consumer facing experiences. We designed workflows with human editors to mitigate semantic or syntax errors. We would like to investigate the automatic detection and mitigation of hallucination errors augmented with fact checkers [39] This work will increase the quality of our output while minimizing the human effort required for editing.

### ACKNOWLEDGMENTS
We would like to thank IBM Corporate Marketing and the MIT-IBM Watson AI Lab for their support during our investigations.